\documentclass[letterpaper, 10 pt, conference]{ieeeconf}
\IEEEoverridecommandlockouts 
\usepackage{cite}
\usepackage{amsmath,amssymb,amsfonts}
\usepackage{algorithmic}
\usepackage{graphicx}
\usepackage{textcomp}
\usepackage{xcolor}
\usepackage{tabularx}
\usepackage{booktabs}
\usepackage{makecell}
\usepackage[caption=false,font=footnotesize]{subfig} 
\usepackage{url}
\def\BibTeX{{\rm B\kern-.05em{\sc i\kern-.025em b}\kern-.08em
    T\kern-.1667em\lower.7ex\hbox{E}\kern-.125emX}}

\begin{document}

\title{\LARGE \bf
KiVi: Kinesthetic-Visuospatial Integration for Dynamic and Safe Egocentric Legged Locomotion}
\author{
Peizhuo Li$^{1*}$, Hongyi Li$^{1,2*}$, Yuxuan Ma$^{1*}$, Linnan Chang$^{1}$, Xinrong Yang$^{1}$, Ruiqi Yu$^{3}$, Shuhao Liao$^{1}$, \\
Yifeng Zhang$^{1}$, Yuhong Cao$^{1\dagger}$, Qiuguo Zhu$^{3}$, Guillaume Sartoretti$^{1}$
\thanks{
$^{*}$ Equal contribution.
$^{\dagger}$ Corresponding author.}
\thanks{$^{1}$ MARMot Lab, National University of Singapore, Singapore.}
\thanks{$^{2}$ Center of X-Mechanics, Zhejiang University, Hangzhou, China.}
\thanks{$^{3}$ Robot and Robot Intelligence Lab, Zhejiang University, Hangzhou, China.}
}

\maketitle

\begin{abstract}
Vision-based locomotion has shown great promise in enabling legged robots to perceive and adapt to complex environments.
However, visual information is inherently fragile, being vulnerable to occlusions, reflections, and lighting changes, which often cause instability in locomotion. 
Inspired by animal sensorimotor integration, we propose KiVi, a \underline{Ki}nesthetic-\underline{Vi}suospatial integration framework, where kinesthetics encodes proprioceptive sensing of body motion and visuospatial reasoning captures visual perception of surrounding terrain.
Specifically, KiVi separates these pathways, leveraging proprioception as a stable backbone while selectively incorporating vision for terrain awareness and obstacle avoidance.
This modality-balanced, yet integrative design, combined with memory-enhanced attention, allows the robot to robustly interpret visual cues while maintaining fallback stability through proprioception.
Extensive experiments show that our method enables quadruped robots to stably traverse diverse terrains and operate reliably in unstructured outdoor environments, remaining robust to out-of-distribution(OOD) visual noise and occlusion unseen during training, thereby highlighting its effectiveness and applicability to real-world legged locomotion. Project Page: \url{https://marmotlab.github.io/kivi-quadruped/}
\end{abstract}

\section{Introduction}

Kinesthetic sense and visuospatial perception constitute two fundamental modalities that allow legged animals to achieve effective locomotion. 
On the one hand, kinesthetic feedback allows animals to maintain balance and adjust body posture with precision.
On the other hand, visual information facilitates navigation through complex terrains and supports pre-emptive avoidance of potential hazards.
For legged robots, these mechanisms remain equally valid, but their integration has been under-studied to date.
Although reinforcement learning controllers for legged robots have demonstrated effective performance without vision ~\cite{lee2020learning, margolis2023walk, hu2019learning, rudin2022learning}, they remain incapable of negotiating terrains that require responding in advance, such as gaps or tall obstacles.
Therefore, to achieve full autonomy for robots in unstructured and dynamic environments, it is necessary to integrate exteroceptive perception into their control frameworks to significantly enhance situational awareness and long-term adaptability~\cite{long2024learninghumanoidlocomotionperceptive}.

However, the addition of visual perception does not always perform as effectively as expected.
Compared to proprioception, which provides a compact but highly informative low-dimensional representation of the robot's state, visual input is sparser in both temporal resolution and relevance to physical interactions, necessitating compression or reformatting via encoders such as CNNs or MLPs~\cite{miki2022learning}.
Additionally, visual sensors are highly susceptible to structured disturbances in real-world scenarios, where reflections and occlusions often lead to misinterpretation of the surrounding terrain. 
These issues compromise the reliability of the overall control strategy.
Within reinforcement learning (RL) pipelines, the inherent disturbances of real-world visual sensors are difficult to model accurately, which can further degrade policy performance through out-of-distribution (OOD) effects~\cite{Yu2022VisualLocomotion}. 

\begin{figure}
    \vspace{0.2cm}
    \centering
    \includegraphics[width=\linewidth]{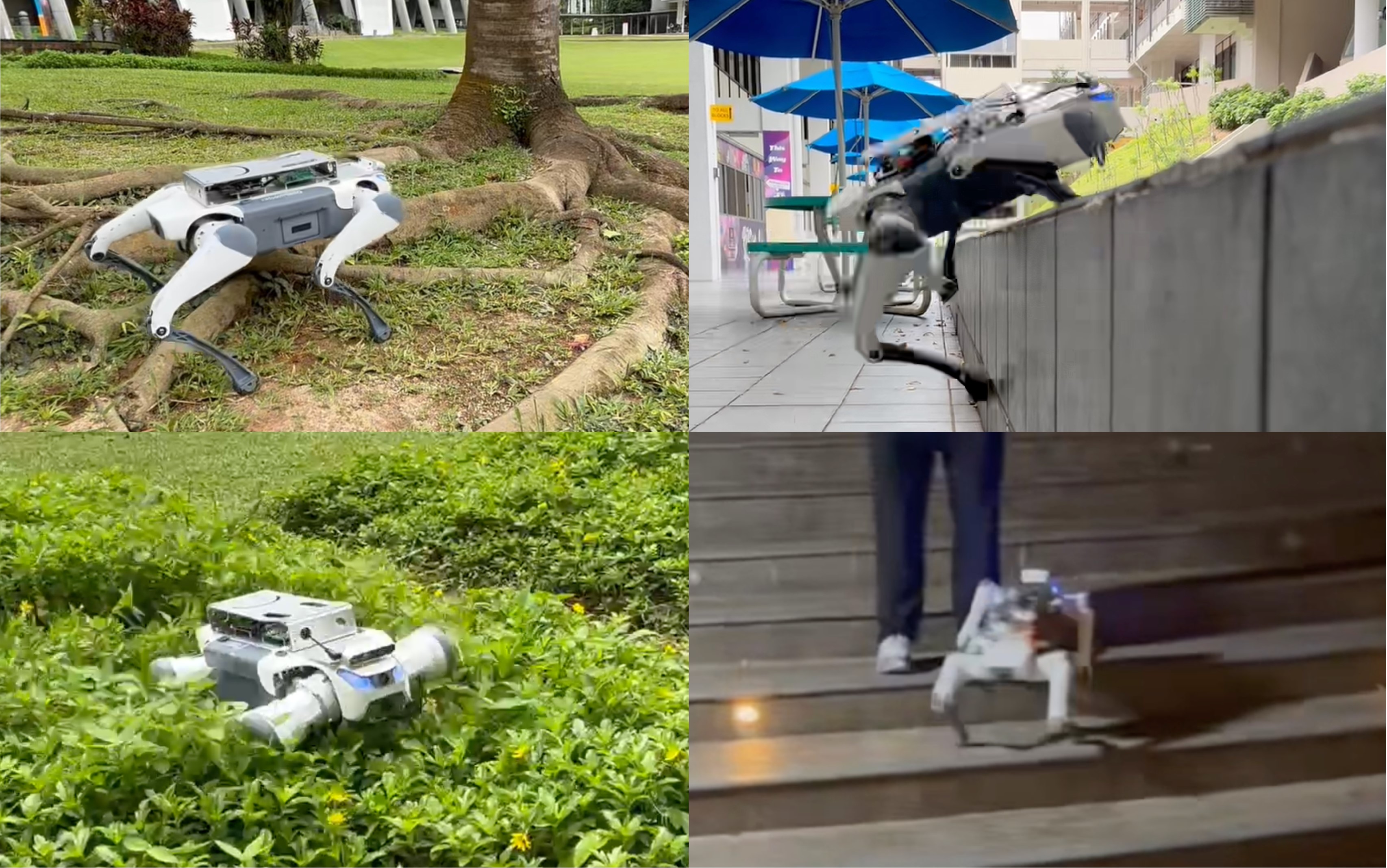}
    \vspace{-0.75cm}
    \caption{Robust locomotion and obstacle avoidance of Deeprobotics Lite3 across diverse terrains and under severe visual disturbances, achieved using our proposed KiVi framework.}
    \vspace{-0.4cm}
    \label{fig:front_page}
\end{figure}

Current vision-based controller pipelines often fuse visual and proprioceptive information, then augment the compressed information with memory mechanisms~\cite{li2024move, miki2022learning}, to finally generate a latent representation that can serve as observation for the policy network.
In these approaches, although attention mechanisms have improved multimodal integration, these methods still suffer from substantial performance degradation.
This is largely due to strong modality entanglement that prevents effective fallback to proprioception when vision becomes unreliable under severe disturbances or degraded signals.
Consequently, robust and reliable operation under such challenging conditions remains difficult to achieve.

In nature, legged animals address this challenge by employing a robust functional division between exteroception and proprioception in complex environments: their movement primarily relies on internal observations to maintain basic stability and coordination~\cite{Moon2021Proprioception}, while vision serves as a complementary source~\cite{Matthis2020VisionWalking}.
When visual input is reliable, they increasingly depend on it for environmental perception to avoid obstacles and maneuver adaptively to traverse complex terrains.
However, when vision is unreliable, e.g. galloping in tall grass, proprioception alone can still support stable locomotion, thereby preventing a complete breakdown of movement when vision is limited ~\cite{Akay2021FeedbackLocomotion}.

Building on this insight, we introduce the \textbf{KiVi} framework, which regulates multimodal information flow through two dedicated modules.
The \textbf{Kinesthetic Module} encodes proprioceptive signals into a compact latent representation that provides a stable backbone for locomotion control.
In parallel, the \textbf{Visuospatial Module} integrates visual observations with proprioceptive context and employs a \textbf{MemTransformer}~\cite{burtsev2021memorytransformer} to enhance temporal memory, enabling the system to reconstruct terrain structures and anticipate upcoming obstacles.
The two latent representations are then concatenated and passed to downstream components, allowing the robot to exploit vision for adaptive terrain traversal and obstacle avoidance while relying on proprioception to sustain robust gait generation.
This separation mechanism ensures that, due to the inherently different signal-to-noise ratios across modalities—where visual observations are typically noisier—the policy naturally prioritizes the more reliable proprioceptive feedback when conflicting information arises, thereby producing more robust behaviors. 
Meanwhile, MemTransformer accelerates training convergence while ensuring that the robot's memory mechanism enables accurate terrain representation.
Extensive experiments demonstrate that KiVi achieves superior performance in diverse real-world scenarios, including stable braking at near-minimum stopping distances, dynamic gap jumping, climbing over high walls, and maintaining central alignment in narrow corridors.
Moreover, the framework exhibits minimal degradation under structured disturbances (e.g., reflective surfaces or dense foliage) and continues to function reliably during complete camera occlusion, establishing a solid foundation for long-term autonomous locomotion in complex and visually challenging environments.

Our main contributions can be summarized as follows:
\begin{itemize}
    \item \textbf{Robust vision-based locomotion framework:} We propose KiVi, a framework that maintains stability under severe disturbances, supports zero-shot sim-to-real transfer on physical robots, and enables near-field obstacle avoidance and complex terrain traversal.

    \item \textbf{Modality-separated design:} By explicitly decoupling proprioceptive and visual pathways, our method relies on proprioception as a reliable backbone, while utilizing vision for adaptive behavior. 
    This design ensures robust policy performance even under OOD conditions with severely corrupted or misleading visual input.

    \item \textbf{Application-oriented functional design:} Experiments demonstrate that our policy achieves robust performance in challenging outdoor environments, autonomously performing tasks such as obstacle avoidance, climbing, and gap jumping, thereby underscoring its effectiveness and applicability for real-world legged locomotion.
\end{itemize}

\section{Related Work}
Legged robots, long recognized for their ability to traverse complex terrains, have attracted sustained research attention for decades~\cite{Hyun2014MITCheetah, Hutter2016ANYmal}. However, their high degrees of freedom and inherent nonlinearity present significant challenges for controller design. Compared to traditional control methods~\cite{di2018dynamic, kim2019highly, ding2019real}, reinforcement learning (RL)-based control policies have shown greater tractability and adaptability, enabling robust locomotion on moderately challenging terrains such as gravel slopes and uneven surfaces~\cite{Peng-RSS-20,rudin2022learning,li2025sata}. Building on these advantages, recent research has increasingly focused on unlocking the full potential of quadrupedal robots, with the goal of achieving stable and dynamic mobility in highly complex and unstructured real-world environments.

\subsection{Proprioceptive Locomotion}
Proprioceptive locomotion constitutes the foundation of RL based locomotion for legged robots. These approaches typically rely exclusively on onboard sensory data such as IMU and joint encoder measurements as policy inputs, and use an actor-critic network architecture to achieve stable and reliable locomotion. In such methods, a key challenge lies in accurately estimating the state of the robot, especially its velocity, when performing complex maneuvers using only onboard sensors~\cite{10801648}. Early approaches introduced privileged observations for critics, allowing the actor–critic framework to evaluate the robot’s state and advantage during training more accurately, making it possible to train robots that can traverse complex terrains under partial observations~\cite{rudin2022learningwalkminutesusing}. To further improve the real-world deployment, the incorporation of historical information to assist in state estimation is proved to be highly effective. For example, ~\cite{lee2020learning, adi9579, kumar2021rma} employed a two-stage teacher-student distillation framework to leverage historical data, allowing real-world deployment in complex terrains including stairs or boulders. Alternatively, other approaches have streamlined this process into a single-stage method. For instance, the DreamWAQ~\cite{nahrendra2023dreamwaq} architecture employs supervised learning to estimate velocity and future states directly from historical data, achieving comparable performance while simplifying the training procedure.

Although these proprioception-based methods are fairly robust due to their relatively low-noise internal sensors, the lack of exteroception makes it more difficult to traverse complex terrains such as gaps,  ditches, or elevated platforms (such as elevated footpaths, large steps, roadside curbs, etc.), preventing them from fully exploiting their inherent structural ability to handle challenging environments.

\subsection{Perception-based Locomotion}
To address the aforementioned limitations, many works have looked into incorporating perception such as vision into RL frameworks. Early works computed elevation maps via traditional methods as policy observations~\cite{miki2022learning} or employed simplified exteroceptive sensing to achieve high-speed locomotion and obstacle avoidance~\cite{he2024agile}. To further unlock the potential of legged platforms, end-to-end learning pipelines have been proposed. Building upon the teacher-student distillation training paradigm, algorithms such as ~\cite{agarwal2022egocentric, zhuang2023parkour, cheng2023extreme, hoeller2023anymal} enabled robots to perform dynamic maneuvers across highly varied terrains. Other methods focused on improvements based on the DreamWAQ~\cite{nahrendra2023dreamwaq} framework. For example, PIE~\cite{luo2024pie} and WMP~\cite{11128762} used supervised learning that integrates historical information and vision to predict the surrounding height map and the next robot observation, achieving single-stage parkour training while significantly reducing training complexity. 

\begin{figure*}[t]
    \centering
    \includegraphics[width=\textwidth]{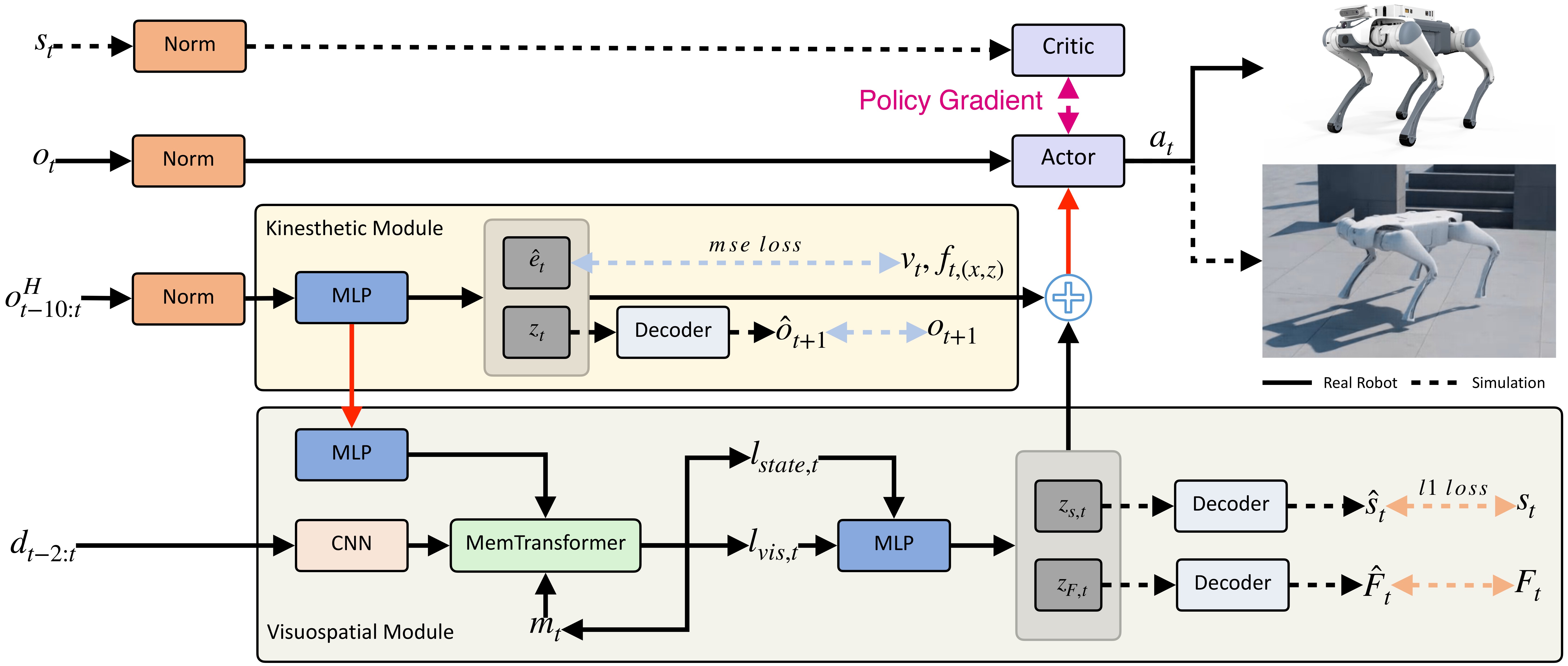}
    \vspace{-0.75cm}
    \caption{Overview of the \textbf{KiVi} framework.
Our bio-inspired dual-branch estimator consists of the \textbf{Kinesthetic Module} (highlighted in yellow) and the \textbf{Visuospatial Module} (highlighted in gray), focusing on proprioceptive information and the integration of visual inputs, respectively.
Solid lines indicate components that are deployed on the real robot, while dashed lines denote parts used only during simulation training.
Red lines represent gradient blocking between modules during training.}
\vspace{-0.4cm}
    \label{fig:framework}
\end{figure*}

However, contrary to expectations, while vision equips the controller with environmental perception and anticipatory planning, it also significantly reduces the robustness of the controller under varying lighting conditions. For depth data to function effectively as an end-to-end input, it requires not only additional domain randomization during simulation training but also sufficiently favorable real-world conditions, such as the absence of severe reflections, occlusions, or poor visibility. In particular, environments with dense vegetation that partially occlude the camera can lead policies to exhibit erratic or overly aggressive actions, thereby limiting their generalizability and practical effectiveness.

\section{Kinesthetic-Visuospatial Integration}
To achieve robust and adaptive locomotion in legged robots, we propose \textbf{KiVi}, a bio-inspired perception–based locomotion framework that emulates key mechanisms by which animals coordinate proprioceptive and visual information. As illustrated in Fig.~\ref{fig:framework}, our framework consists of two main components: an asymmetric actor–critic architecture for efficient training and a bio-inspired dual-branch estimator that assigns dedicated pathways to proprioceptive and visual inputs. The estimator produces compact latent representations that prioritize internal dynamics while selectively incorporating complementary visual cues, which are used by the actor to generate control actions, while the critic is augmented with privileged information during training. This structured design safeguards the quality of learned representations, thereby enhancing policy stability under challenging perceptual conditions, similar to how animals rely on proprioception as a stable backbone while incorporating visual feedback to navigate complex terrains.

\subsection{Actor–Critic Architecture}
We employ an \textbf{Asymmetric Actor–Critic Architecture}, in which the critic is granted access to privileged information, whereas the actor operates exclusively on proprioceptive observations together with the kinesthetic and visuospatial latent representations provided by the estimator. The asymmetric design improves value estimation and stabilizes policy updates, thereby enhancing both sample efficiency and robustness, while maintaining a deployable observation-only actor policy.
By unifying perception and control within a single-stage framework, our approach avoids the error accumulation inherent to teacher–student frameworks, while directly optimizing task performance under encoder observations. 
We optimize the policy using proximal policy optimization (PPO)~\cite{schulman2017proximal}, resulting in an efficient and stable training process.

\subsubsection{Actor Network}
Our actor network is provided with a 45-dimensional proprioceptive observation $\mathbf{o}_t$, a 31-dimensional Kinesthetic latent embedding, and a 20-dimensional Visuospatial latent embedding. The proprioceptive observation $\mathbf{o}_t$ is defined as:
\begin{equation}
\mathbf{o}_t =
\begin{bmatrix}
\boldsymbol{\omega}_t & \mathbf{g_t} & \mathbf{c_t} & \boldsymbol{\theta}_t & \dot{\boldsymbol{\theta}}_t & \mathbf{a}_{t-1}
\end{bmatrix}^T,
\end{equation}
where $\boldsymbol{\omega}_t$ denotes the body angular velocity, $\mathbf{g_t}$ is the gravity direction vector expressed in the body frame, and $\mathbf{c_t}$ represents the velocity command. $\boldsymbol{\theta}_t$ and $\dot{\boldsymbol{\theta}}_t$ correspond to joint positions and joint velocities, respectively, while $\mathbf{a}_{t-1}$ denotes the action produced by the actor network in the previous timestep.
\subsubsection{Critic Network}
The critic network leverages privileged observations obtained in the simulation environment to accurately estimate the state value, thereby facilitating the update of actor parameters. The input to the critic network at time step $t$ is defined as:
\begin{equation}
\mathbf{s}_t =
\begin{bmatrix}
\mathbf{v}_t & \mathbf{o}_t & \mathbf{f}^{xy,z}_t & \mathbf{m}^{b}_t
\end{bmatrix}^T,
\end{equation}
where $\mathbf{v}_t$ denotes the base velocity, $\mathbf{f}^{xy,z}_t$ represents eight contact forces of the four feet in both $x$–$y$ plane and $z$ direction, and $\mathbf{m}^{b}_t$ represents the local height scans around the base.
\subsubsection{Reward}
\begin{table}[t]
\caption{Reward Components and Weights ($dt = 0.02$).}
\vspace{-0.3cm}
\centering
\renewcommand{\arraystretch}{1.4} 
\begin{tabular}{lcc}
\toprule
\textbf{Reward Terms} & \textbf{Equation ($r_i$)} & \textbf{Weight ($w_i$)} \\
\midrule
\multicolumn{3}{l}{\textbf{Locomotion Objectives}} \\
$r_{\text{tracking},xy}$ & $\phi \!\left(v_{xy} - v_{xy}^{cmd}\right)$ & $3dt$ \\
$r_{\text{tracking},yaw}$ & $\phi \!\left(\omega_{yaw} - \omega_{yaw}^{cmd}\right)$ & $1.5dt$ \\
\midrule
\multicolumn{3}{l}{\textbf{Behavioral Constraints and Penalties}} \\
$r_{\text{velocity},z}$   & $v_z^2$ & $-0.1dt$ \\
$r_{\text{ang. vel.},xy}$ & $\boldsymbol{\omega}_{xy}^2$ & $-0.05dt$ \\
$r_{\text{joint acc.}}$ & $\ddot{\boldsymbol{\theta}}^2$ & $-2.5 \times 10^{-7}dt$ \\
$r_{\text{joint power}}$ & $|\boldsymbol{\tau}||\dot{\boldsymbol{\theta}}|$ & $-2 \times 10^{-5}dt$ \\
$r_{\text{joint torque}}$ & $\boldsymbol{\tau}^2$ & $-1 \times 10^{-5}dt$ \\
$r_{\text{power dist.}}$ & $\operatorname{var}| \boldsymbol{\tau} \cdot \dot{\boldsymbol{\theta}}|$ & $-2 \times 10^{-7}dt$ \\
$r_{\text{collision}}$ & $-n_{\text{collision}}$ & $-1dt$ \\
$r_{\text{action rate}}$ & $(\mathbf{a}_t - \mathbf{a}_{t-1})^2$ & $-0.01dt$ \\
$r_{\text{smoothness}}$ & $(\mathbf{a}_t - 2\mathbf{a}_{t-1} + \mathbf{a}_{t-2})^2$ & $-0.01dt$ \\
\bottomrule
\end{tabular}
\vspace{-0.4cm}
\label{tab:rew}
\end{table}
To demonstrate the effectiveness of our framework, we adopt a commonly used reward design and parameter settings for legged locomotion control tasks~\cite{nahrendra2023dreamwaq,kumar2021rma} without extensive task-specific customization or fine-tuning. The details of each reward terms are comprehensively summarized in Table~\ref{tab:rew}.

\subsubsection{Action}
Our framework employs low-level position control, where the actor network outputs a 12-dimensional vector $\mathbf{a}_t$ at each timestep, which is added to the predefined default joint positions $\boldsymbol{\theta}^{\text{default}}$ to obtain the desired joint positions for all joints $\boldsymbol{\theta}^{\text{target}}_t$, as shown below
\begin{equation}
\boldsymbol{\theta}^{\text{target}}_t = \boldsymbol{\theta}^{\text{default}} + \mathbf{a}_t.
\end{equation}
The desired joint positions are then fed into a low-level PD controller to compute the target joint torques $\boldsymbol{\tau}_t$, which are defined as:
\begin{equation}
\boldsymbol{\tau}_t = K_p \left(\boldsymbol{\theta}^{\text{target}}_t - \boldsymbol{\theta}_t \right) 
- K_d \, \dot{\boldsymbol{\theta}}_t,
\end{equation}
where the stiffness $K_p$ and the damping $K_d$ are set to $30.0$ and $1.0$, respectively.

\subsection{Bio-inspired Dual-branch Estimator}
To enable robust multimodal perception, we design a \textbf{bio-inspired dual-branch estimator} that emulates the complementary roles of proprioception and vision in animal locomotion. Biological systems rely on proprioception as a dependable backbone for balance and coordination, while selectively incorporating visual information to perceive external obstacles and terrain  in a dynamic and adaptive manner. Inspired by this principle, our estimator separates sensory processing into two dedicated pathways.

The \textbf{Kinesthetic Module} processes a short sequence of history proprioceptive observations to infer internal dynamics, producing a latent representation that captures the current and the next state of the robot. In parallel, our \textbf{Visuospatial Module} integrates egocentric depth images with temporal context to generate a compact terrain embedding, representing surrounding elevations and local foothold height scan.

The two latent representations are then concatenated and passed to the actor network, providing a structured and task-aligned perceptual input. This design reduces the difficulty of representation learning, mitigates the impact of visual disturbances by relying on solid proprioception, and enhances stability and adaptability across diverse environments.

\subsubsection{Kinesthetic Module}
Our Kinesthetic Module leverages short histories of proprioceptive observations $\boldsymbol{o}_{t-10:t}^H$ to infer both explicit and implicit dynamic representations of the robot. The explicit outputs $\hat{\boldsymbol{e}}_t\in \mathbb{R}^{11}$ include the estimated base linear velocity and the estimated foot contact forces in the $x$ and $z$ directions. These quantities are supervised with an MSE loss against privileged ground truth values ${\boldsymbol{e}}_t$ in simulation, providing reliable estimation of the robot’s state and contact conditions that are otherwise difficult to measure directly with onboard sensors. In parallel, our module generates an implicit latent vector $\boldsymbol{z}_t\in \mathbb{R}^{20}$ that predicts the next-step observation $\hat{\boldsymbol{o}}_{t+1}$ in a VAE-style manner, encouraging the network to capture latent dynamic priors beyond the explicitly supervised targets.

By jointly capturing the explicit estimate $\hat{\boldsymbol{e}}_t$ and the implicit prediction $\boldsymbol{z}_t$, our Kinesthetic Module provides the policy with stable and temporally consistent internal state representations. This design not only strengthens short-term dynamics modeling but also enables robust self-assessment under noisy proprioceptive readings or rapidly changing contact conditions, thereby enhancing both the robustness and generalization of the learned locomotion policy.

\subsubsection{Visuospatial Module}
Our Visuospatial Module focuses on fusing visual and proprioceptive information to infer key terrain-related features, such as surrounding elevations and foot-level height map, which cannot be directly derived from proprioception alone. This process is challenging as the robot cannot always directly observe the ground beneath its feet, making effective multimodal integration and temporal memory essential for reliable terrain understanding.

To address these challenges, our module takes as input the past 10 steps of proprioceptive observations $\boldsymbol{o}_{t-10:t}^H$ and the past 2 frames of egocentric depth images $\boldsymbol{d}_{t-2:t}$. The proprioceptive sequence is normalized and encoded by an MLP into a single token, while the depth sequence is processed by a convolutional neural network (CNN) to produce 16 visual tokens.
These tokens are then fused within a memory-augmented Transformer (\textbf{memTransformer}). At each timestep, the Transformer receives a total of 20 tokens: 1 proprioceptive token $l_{state,t} \in R^{32}$, 16 visual tokens $l_{vis,t} \in R^{1 6\times 32}$, and 3 memory tokens $m_{t} \in R^{3 \times 32}$ inherited from the previous timestep. Our memTransformer consists of two stacked Transformer layers with a single attention head (i.e., $N_{\text{head}} = 1$) and a feedforward dimension of $128$. 
After attention processing, the 20 output tokens are handled asymmetrically. The updated proprioceptive token is directly propagated to downstream policy layers. The 16 visual tokens are average-pooled into a single visual embedding before being forwarded. In contrast, the 3 memory tokens are not passed to the policy head; instead, they are cached and reused as memory inputs at the next timestep. This design allows the model to maintain a compact, content-addressable memory buffer while keeping the policy input dimension fixed.
Compared with recurrent models such as GRUs, this mechanism provides a more flexible balance between parameter efficiency and representational capacity. A small number of persistent memory tokens can selectively accumulate long-horizon contextual information without enforcing strict sequential compression as in recurrent hidden states. 
Moreover, by avoiding recursive hidden-state updates and reducing long backpropagation chains through time, our memTransformer improves training stability and sample efficiency, leading to faster convergence in long-horizon tasks.

Through this mechanism, our module implicitly predicts the global surrounding elevation $\hat{s}_t$ and the local terrain heights around each foot $\hat{F}_t$, and produces latent embeddings $z_{s,t}\in \mathbb{R}^{12}$ and $z_{F,t}\in \mathbb{R}^{8}$ that encode both coarse terrain structure and fine-grained foothold surroundings. The predictions $\hat{s}_t$ and $\hat{F}_t$ are supervised with a stricter $L_1$ loss against privileged terrain information available in simulation. By jointly leveraging visual and proprioceptive cues and maintaining temporal memory, our Visuospatial Module provides stable and informative terrain embeddings, complementing the Kinesthetic Module and enabling reliable locomotion control even under partial or noisy visual observations.

The outputs of our Kinesthetic and Visuospatial Modules, namely $\hat{\boldsymbol{e}}_t$, $\boldsymbol{z}_t$, $\boldsymbol{z}_{s,t}$, and $\boldsymbol{z}_{f,t}$, are concatenated into a single vector and fed into the actor network. The overall loss of the estimator is defined as:
\begin{align}
\mathcal{L} = \;&
D_{\mathrm{KL}}\!\left(q(\boldsymbol{z}_t \mid \boldsymbol{o}^H_{t-10:t}, \boldsymbol{d}_{t-2:t}) \,\|\, p(\boldsymbol{z}_t)\right) 
+ \mathrm{MSE}(\hat{\boldsymbol{e}}_t, \boldsymbol{e}_t) \notag \\
&+ \mathrm{MSE}(\hat{\boldsymbol{o}}_{t+1}, \boldsymbol{o}_{t+1})
+ \ell_1(\hat{\boldsymbol{s}}_t, \boldsymbol{s}_t) 
+ \ell_1(\hat{\boldsymbol{F}}_t, \boldsymbol{F}_t),
\end{align}
where $q(\boldsymbol{z}_t \mid \boldsymbol{o}^{H}_{t-10:t}, \boldsymbol{d}_{t-2:t})$ denotes the posterior distribution of $\boldsymbol{z}_t$ conditioned on $\boldsymbol{o}^{H}_{t-10:t}$ and $\boldsymbol{d}_{t-2:t}$, while $p(\boldsymbol{z}_t)$ represents the prior distribution of $\boldsymbol{z}_t$ which is typically parameterized as a standard normal distribution.

\begin{table}[t]
\caption{Domain Randomization Ranges.}
\vspace{-0.3cm}
\centering
\begin{tabular}{lcc}
\toprule
\textbf{Parameter} & \textbf{Randomization range} & \textbf{Unit} \\
\midrule
Payload & $[-1, 3]$ & kg \\
$K_p$ factor & $[0.9, 1.1]$ & Nm/rad \\
$K_d$ factor & $[0.9, 1.1]$ & Nms/rad \\
Center of mass shift & $[-50, 50]$ & mm \\
Static friction coefficient & $[0.5, 1.25]$ & - \\
Dynamic friction coefficient & $[0.3, 1.1]$ & - \\
Initial joint positions & $[0.5, 1.5]$ & rad \\
System delay & $[0, 20]$ & ms \\
Camera shaking & $[-2, 2]$ & deg \\
\bottomrule
\end{tabular}
\vspace{-0.4cm}
\label{tab:randomization}
\end{table}

\subsection{Training details}
\subsubsection{Simulation Platform}
Leveraging NVIDIA Isaac Sim and Isaac Lab, we train 4096 agents in parallel in simulation; our final policy can be directly deployed on Deeprobotics Lite3 robot. Training converges after approximately 12,000 iterations, about 8 hours on a single NVIDIA RTX 4090.

\subsubsection{Training Terrain and Curriculum}
\begin{figure}[t]
    \centering
    \includegraphics[width=1.0\columnwidth]{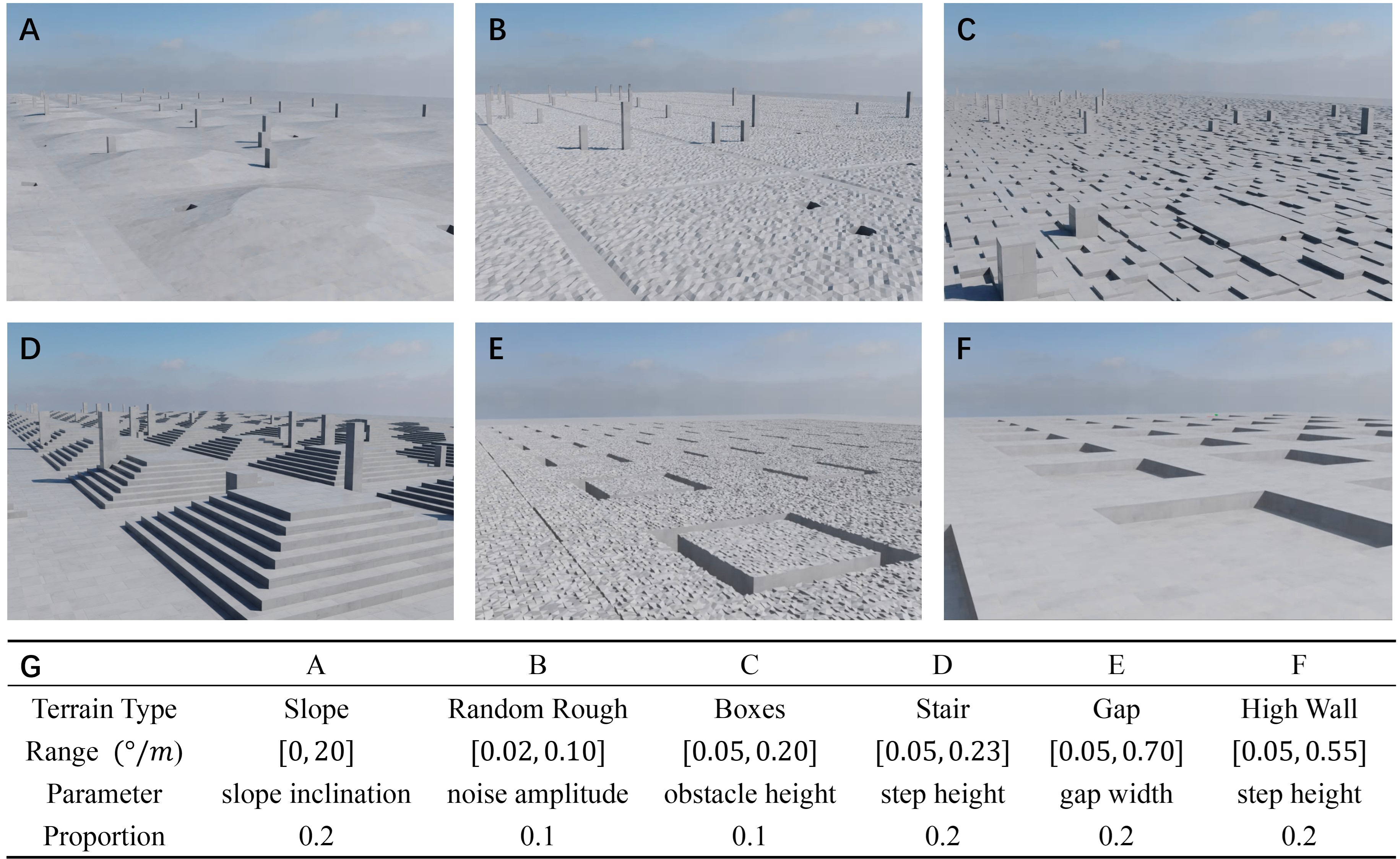}
    \vspace{-0.75cm}
    \caption{Simulated terrain types A–F used during training, each representing a distinct terrain challenge. As the training difficulty increases, each terrain randomly generates 0–5 obstacles. Figure G lists the control parameters and their respective ranges for each terrain type, which are used to procedurally generate diverse terrain instances.}
    \label{fig:terrain}
    \vspace{-0.4cm}
\end{figure}

In simulation, we train the policy simultaneously on six types of terrains: stairs, platforms, random rough, slope, gaps, and high walls. Among them, stairs, platforms, random rough and slope are native terrains provided by Isaac Lab, while gaps and high walls are custom-designed terrains.
To prevent policy collapse in the early stages of training due to overly challenging terrains, we adopt a curriculum~\cite{rudin2022learning} that progressively increases difficulty as the policy improves. The specific terrain and the curriculum settings are shown in Fig.~\ref{fig:terrain}.

\subsubsection{Domain Randomization}
We also use Domain Randomization to enhance policy robustness and facilitate smooth sim-to-real transfer. Our system design ensures that strong performance can be achieved by relying only on standard domain randomization techniques. On the proprioceptive side, we follow a common setting that includes the randomization of the payload, the center of mass, the PD gains ($K_p$, $K_d$) and the ground friction coefficient. On the visual side, aside from standard sensor noise, we introduce camera shaking to randomize camera position, simulating sensor vibrations during deployment. The specific randomization settings are summarized in Table~\ref{tab:randomization}.

\section{Experiments}

To systematically evaluate the robustness of our \textbf{KiVi} framework, we conduct comparative studies, including both ablation experiments and comparisons against representative baselines. These evaluations comprehensively cover algorithmic and functional aspects, providing a holistic understanding of system performance:

\begin{itemize}
    \item \textbf{KiVi w/o Kin.}: A variant where the velocity $v_t$ and the predicted observation $\hat{o}_{t+1}$ are produced from a fused visual–proprioceptive representation, rather than through explicitly separated pathways. This configuration is structurally similar to the representation design adopted in PIE~\cite{luo2024pie}, and serves as a controlled baseline to evaluate the contribution of modality separation to robustness.
    
    \item \textbf{KiVi w/o Memory}: A variant where the MemTransformer is replaced by a standard Transformer, to study the effect of temporal memory in long-horizon tasks.

    \item \textbf{KiVi GRU}: A variant in which the MemTransformer is replaced by a standard Transformer followed by a GRU module. This configuration is designed to assess the effectiveness of the MemTransformer architecture compared to a conventional Transformer+GRU temporal modeling scheme.
    
    \item \textbf{Himloco~\cite{long2023hybrid}}: A blind locomotion policy relying solely on proprioception, which we include to highlight the role of vision in locomotion.
\end{itemize}

All experiments were conducted on the \textbf{DeepRobotics Lite3} quadruped equipped with a RealSense D435i depth camera. Image preprocessing and policy inference were run onboard on a Jetson Orin NX Super. The system operated with depth acquisition at $10\,\text{Hz}$, policy inference at $50\,\text{Hz}$, and a low-level PD controller at $200\,\text{Hz}$.

\subsection{Simulation Experiments}
We first evaluate the robustness under severe visual disturbances in simulation. \textbf{KiVi}, \textbf{KiVi w/o Kin.}, and \textbf{Himloco} were deployed on random rough terrains with corrupted depth inputs, including high-intensity Gaussian noise, large random occlusions, and severe camera jitter. These disturbances are considerably stronger than those encountered during training.

\begin{figure}[t]
    \centering
    \includegraphics[width=\linewidth]{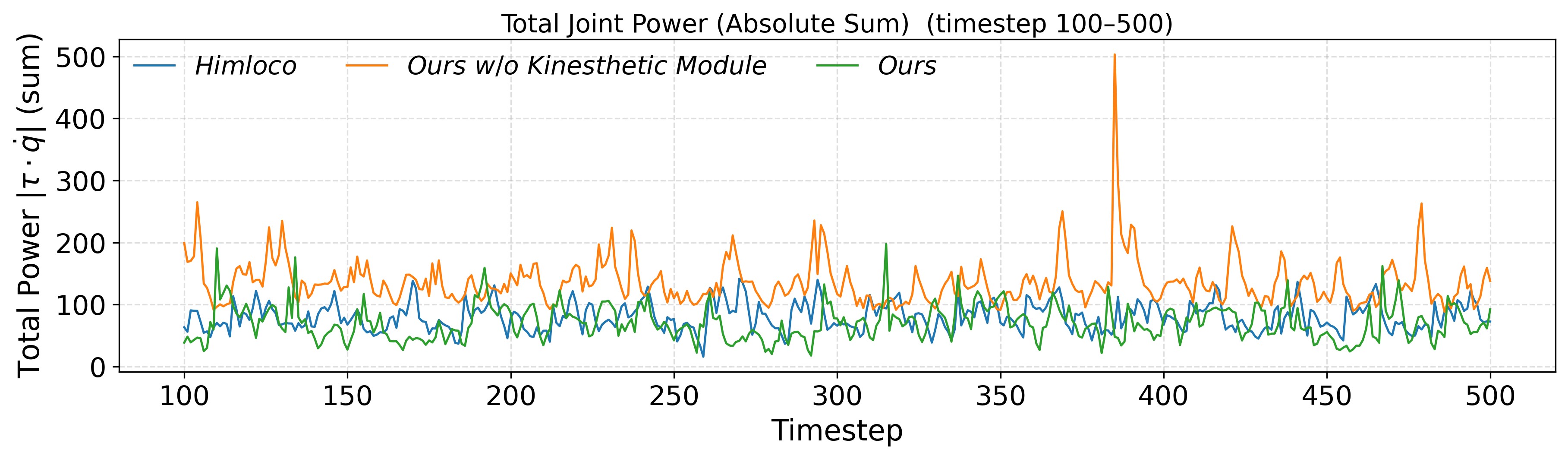}\\[-0.2em]
    \includegraphics[width=\linewidth]{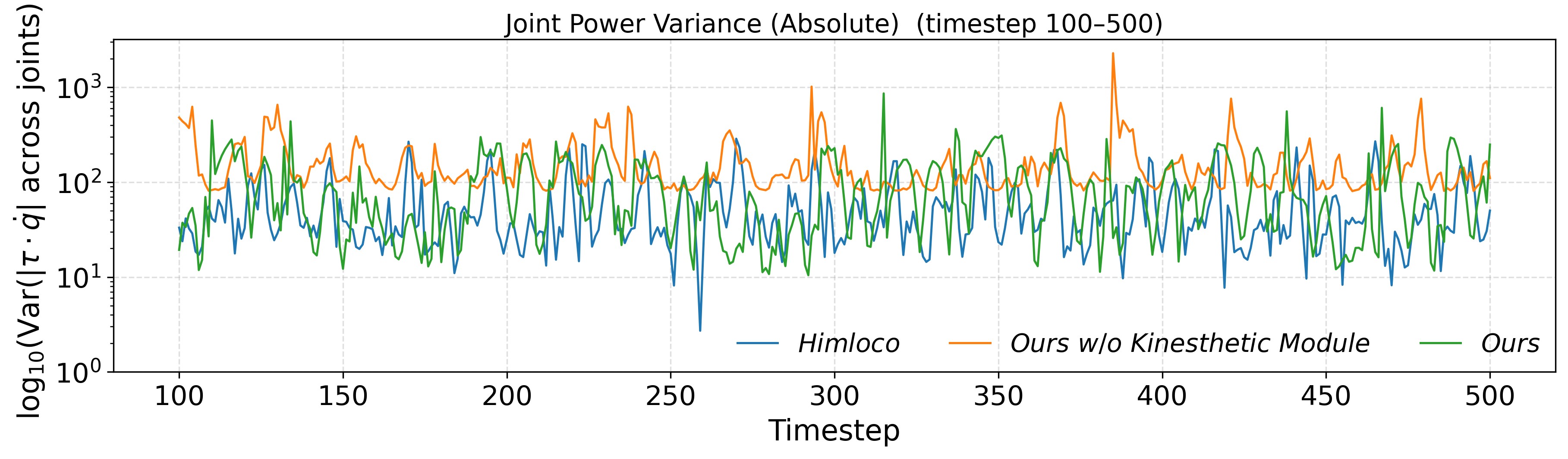}
    \vspace{-0.75cm}
    \caption{Total joint power and power variance across all joints for KiVi, KiVi w/o Kin., and Himloco on simulated rough terrains under severe visual disturbances.}
    \label{fig:jointpower}
\end{figure}

During the experiments, we recorded both the total motor power of joints and the variance of the power distribution across joints (Fig.~\ref{fig:jointpower}). As expected, the blind policy \textbf{Himloco} was unaffected by visual corruption, showing consistently low mean power and minimal variance, indicating well-balanced actuation. In contrast, \textbf{KiVi w/o Kin.} exhibited large power fluctuations and sharp peaks, reflecting severe joint oscillations with high torques and velocities. This unstable behavior leads to higher energy consumption, motor overheating, and increased mechanical wear. In comparison, the full \textbf{KiVi} framework maintained stable operation with energy consumption close to Himloco, demonstrating its robustness and suitability for long-term deployment in disturbed environments.

\subsection{Hardware Experiments}
\begin{figure}[t]
    \centering
    \includegraphics[width=0.95\columnwidth]{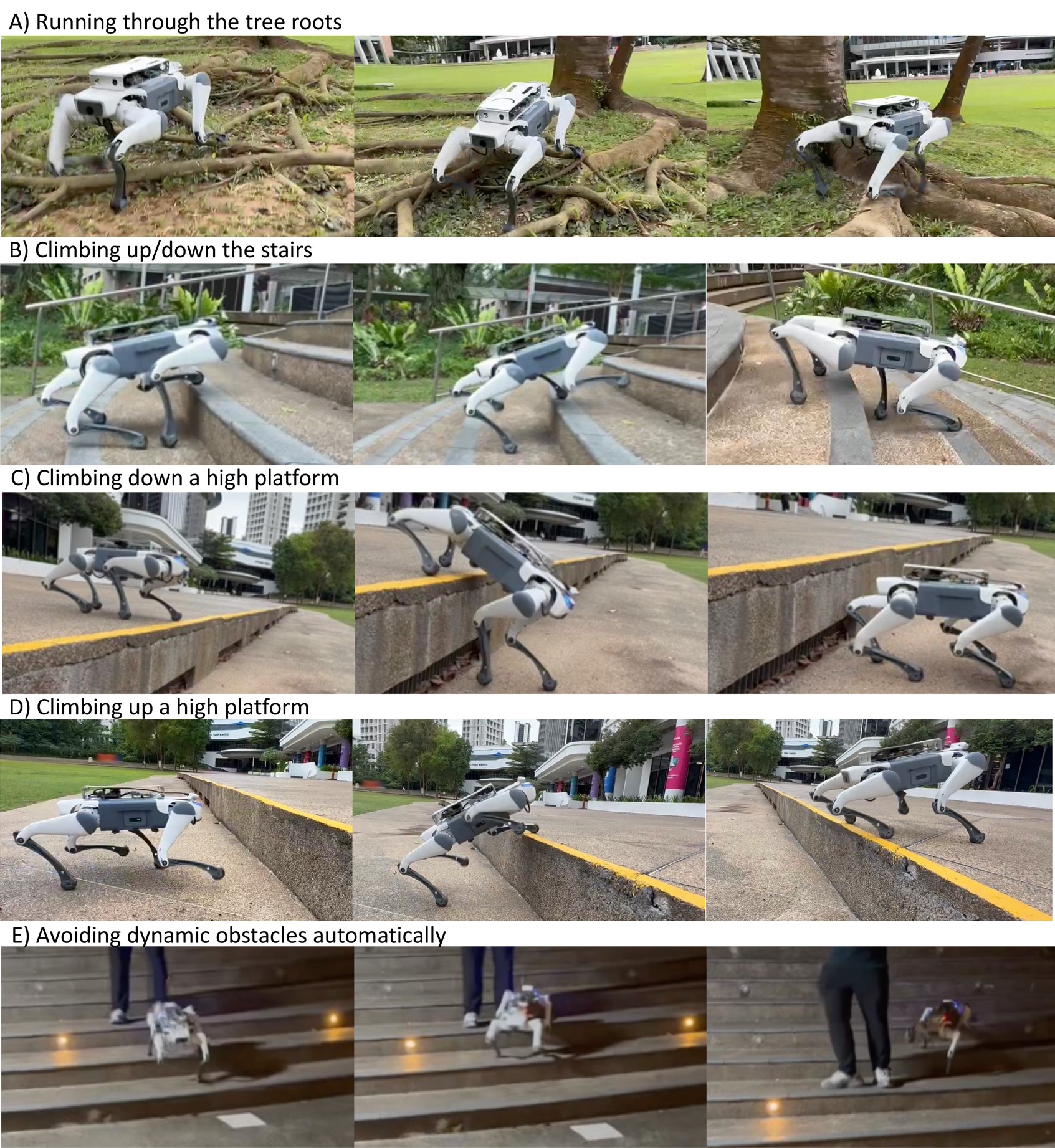}
    \vspace{-0.35cm}
    \caption{Outdoor hardware experiments under low visual disturbances, including tree roots, staircases, elevated platforms, and dynamic pedestrians. With only a constant forward velocity command of $[1.0, 0, 0]$, the robot traversed all terrains and avoided obstacles.}
    \label{fig:exp-low_vis_noise}
\end{figure}

We further evaluated all methods on the robot in outdoor scenarios (Fig.~\ref{fig:exp-low_vis_noise} and Fig.~\ref{fig:exp-tall_grass_covered}). The experimental results  are summarized in Table~\ref{tab:test_result}.

\begin{table}[t]
\caption{Performance comparison on-robot between our method and the baselines across robustness tests (5 trials per test).}
\vspace{-0.2cm}
\centering
\resizebox{0.95\columnwidth}{!}{
\begin{tabular}{lcccc}
\hline
& \makecell{High \\ Platform} 
& \makecell{Obstacle \\ Avoidance} 
& \makecell{Tall \\ Grass} 
& \makecell{Block \\ Camera} \\ 
\hline
\textbf{KiVi} & \textbf{5/5} & \textbf{5/5} & \textbf{5/5} & \textbf{5/5} \\
KiVi w/o Kin. (PIE) & 4/5 & \textbf{5/5} & 3/5 & 0/5 \\
KiVi GRU & 4/5 & \textbf{5/5} & 4/5 & 3/5 \\
KiVi GRU (24000 iters) & \textbf{5/5} & \textbf{5/5} & \textbf{5/5} & \textbf{5/5} \\
KiVi w/o Memory & 3/5 & 2/5 & 2/5 & 0/5 \\
Himloco & 0/5 & 0/5 & \textbf{5/5} & / \\
\hline
\end{tabular}
}
\vspace{-0.3cm}
\label{tab:test_result}
\end{table}

\subsubsection{Terrain Traversability}

In comparative studies, \textbf{KiVi} and \textbf{KiVi w/o Kin.} showed nearly identical performance: both traversed uneven terrains and avoided pedestrians at near-limit distances ($\sim\!3\,\text{cm}$), comparable to \textbf{ABS}~\cite{he2024agile} on flat ground, while exhibiting more stable control on stairs. 

\textbf{KiVi GRU}, which replaces the memTransformer with a Transformer+GRU structure, achieved comparable success rates in obstacle avoidance but showed slightly reduced robustness on high platforms and tall grass under the same training budget. Notably, when trained for extended iterations (24000 iterations), KiVi GRU eventually reached performance parity with \textbf{KiVi}. This suggests that while recurrent compression can ultimately capture sufficient temporal context, our proposed memory-token mechanism facilitates faster convergence and more efficient temporal credit assignment.

\textbf{KiVi w/o Memory}, however, lacks temporal context, which results in more conservative behaviors and reduced stability. \textbf{Himloco}, which does not use vision, responds only after physical contact occurs. As a result, it frequently stumbles at the edge of platforms or collides with obstacles. It also fails to traverse challenging terrains such as gaps and high steps. These results highlight the importance of both multimodal fusion and temporal memory for safe operation in dynamic outdoor environments.

\subsubsection{Visual Robustness}

\begin{figure}[t]
    \centering
    \includegraphics[width=\columnwidth]{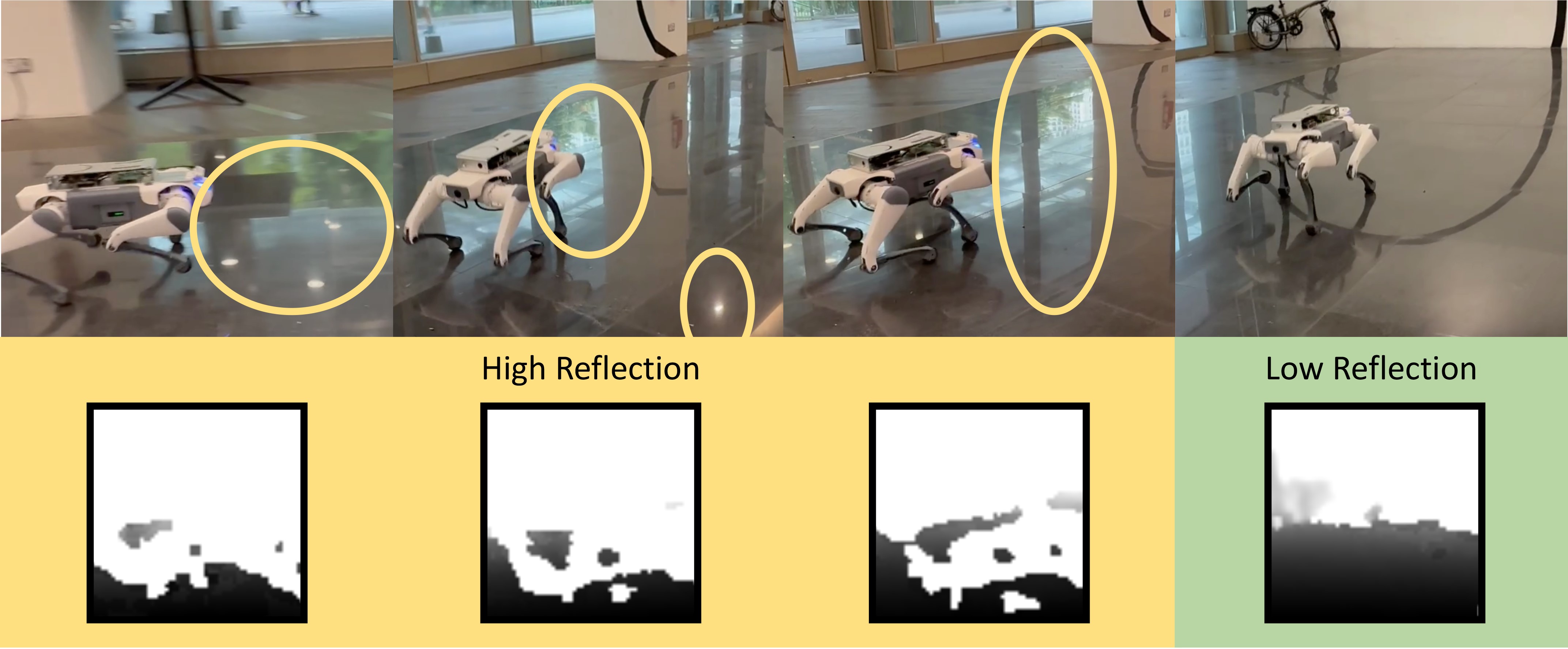}
    \vspace{-0.75cm}
    \caption{Locomotion on reflective surfaces. Strong reflections (yellow) distort the depth image, causing the sensor to misinterpret the ground as distant empty space, while the green region indicates weaker artifacts. Despite these distortions, KiVi maintains stable locomotion.}
    \label{fig:exp-reflection}
    \vspace{-0.4cm}
\end{figure}

\begin{figure}[t]
    \centering
    \includegraphics[width=\columnwidth]{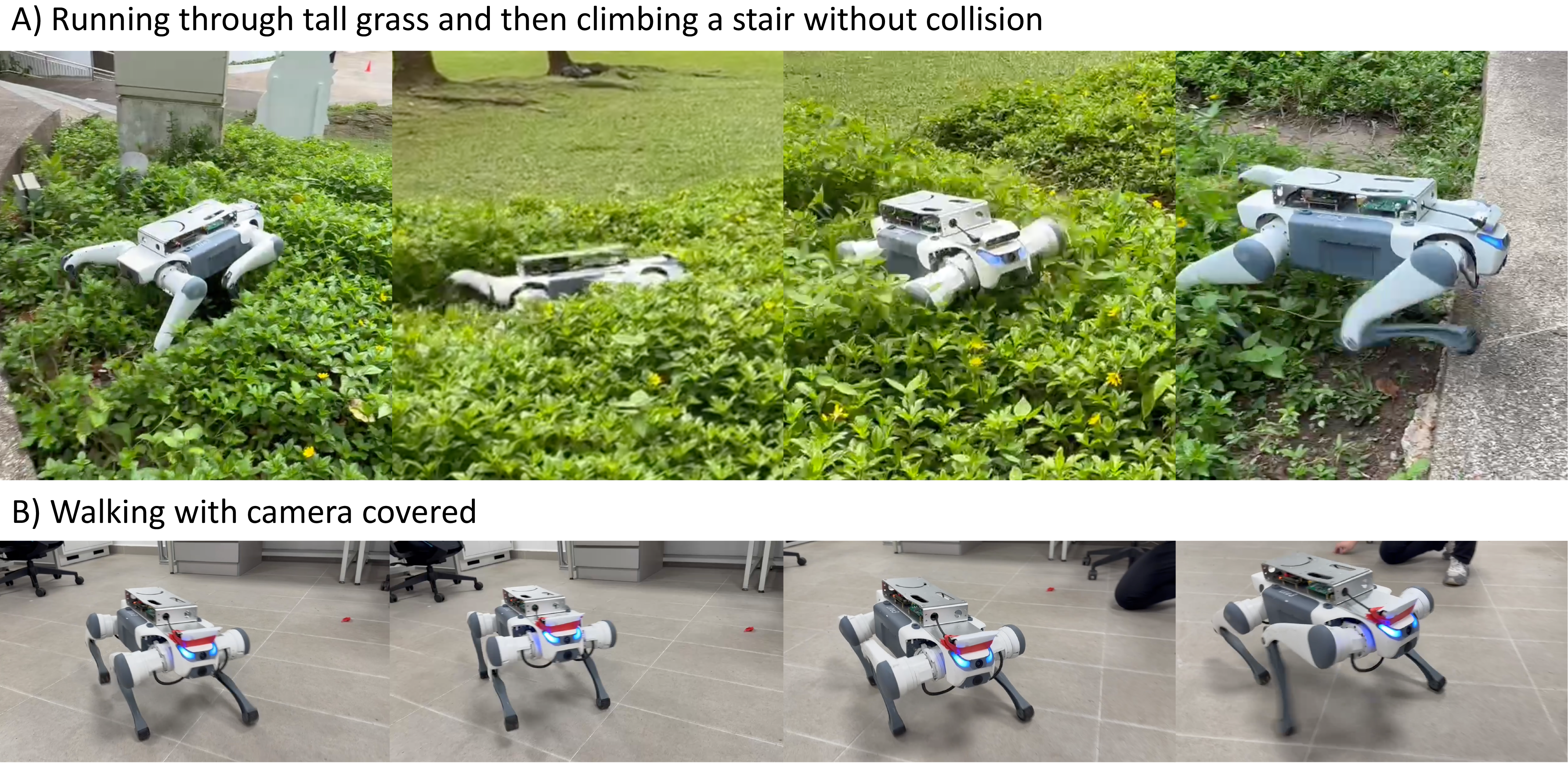}
    \vspace{-0.75cm}
    \caption{Performance of KiVi under strong visual disturbances. 
    (\textbf{A}) Traversing tall grass and adapting gait to climb a platform. 
    (\textbf{B}) Camera fully covered, where the policy seamlessly falls back to proprioception for stable locomotion.}
    \label{fig:exp-tall_grass_covered}
    \vspace{-0.4cm}
\end{figure}

We then tested environments with strong visual disturbances, including abrupt illumination changes, tall grass, reflective surfaces, and complete camera occlusion. Random illumination noise had minor overall impact, but structured disturbances, such as tall grass, provided persistent, misleading semantic cues that directly conflicted with proprioceptive feedback, creating challenging OOD scenarios unseen in training. Unlike random noise, which is largely filtered in network representations, structured noise forced inconsistent interpretations of the terrain, greatly increasing locomotion difficulty.

In these settings, \textbf{KiVi} consistently completed all tasks with only slight reductions in gait stability. For example, in tall grass (Fig.~\ref{fig:exp-tall_grass_covered}A), the robot exited the disturbed area and immediately adjusted its gait to climb onto an elevated platform. Finally, when vision was completely occluded (Fig.~\ref{fig:exp-tall_grass_covered}B), the policy gracefully fell back to proprioception, generating conservative yet stable motions resembling blind locomotion. This demonstrates the value of modality separation: proprioception provides a reliable backbone for safe fallback, even when vision becomes entirely unreliable/unavailable. By contrast, \textbf{KiVi w/o Kin.} suffered from unstable behaviors due to entangled representations, occasionally producing abrupt pitching. We believe these findings highlight that explicit modality separation not only enhances robustness to corrupted vision but also enables graceful degradation under extreme conditions, whereas mixed representations remain vulnerable to OOD conflicts.

\section{Conclusion}

In this work, we introduced \textbf{KiVi}, a framework for robust quadruped locomotion in visually challenging environments. 
By explicitly separating proprioceptive and visual pathways and enhancing multi-modal fusion with a memory-augmented transformer, KiVi maintains stable locomotion by relying on proprioception as a backbone, while selectively exploiting visual cues to enhance terrain awareness and anticipatory obstacle avoidance whenever visual perception is reliable.
Our results show that this design achieves strong resilience even under unstructured visual noise, such as tall grass, reflective surfaces, and camera occlusion, where fused-latent or vision-dominant baselines often fail.
At the same time, our framework exhibits a graceful degradation property: when vision becomes unreliable, control naturally falls back to proprioception without catastrophic failure. 

Despite these advantages, KiVi has certain limitations.
The current memory module captures only short-term temporal information, which may be insufficient for tasks requiring longer persistence; for instance, when the robot pauses in front of an elevated platform, the memory may decay and misinterpret the obstacle.
Extending the memory mechanism or integrating hierarchical reasoning could address this issue.
Looking ahead, we believe that KiVi’s core principles, including modality separation, conditional integration of visual input, and memory-enhanced fusion, provide a solid foundation for robust multimodal control, with potential to extend beyond remote controlled locomotion to long-term autonomous deployment in real-world applications.

\section*{ACKNOWLEDGMENT}

We used ChatGPT to assist in the linguistic refinement of the Introduction and Conclusion sections. The generated text was critically reviewed and revised by the authors to ensure alignment with the research findings and academic standards.

We thank Dr. Chao Li of DEEP Robotics for his valuable assistance with the real-world experiments and for providing hardware support.

This work was supported in part by the Singapore Ministry of Education (MOE), the National University of Singapore under its Robotics Grand Challenge, the “Leading Goose” R\&D Program of Zhejiang under Grant 2023C01177, the National Key R\&D Program of China under Grant 2022YFB4701502, and the 2035 Key Technological Innovation Program of Ningbo City under Grant 2024Z300.

\bibliographystyle{IEEEtran}
\bibliography{references}
\end{document}